\documentclass{article} 
\usepackage{iclr2026_conference,times}

\usepackage{amsmath,amsfonts,bm}

\def\eqref#1{equation~\ref{#1}}

\def\1{\bm{1}}

\DeclareMathAlphabet{\mathsfit}{\encodingdefault}{\sfdefault}{m}{sl}
\SetMathAlphabet{\mathsfit}{bold}{\encodingdefault}{\sfdefault}{bx}{n}

\usepackage{hyperref}
\usepackage{url}
\usepackage[most]{tcolorbox}
\usepackage{booktabs}   
\usepackage{multirow}   
\usepackage{booktabs}
\usepackage{multirow}
\usepackage{xcolor}
\usepackage{wrapfig}
\usepackage{bm}
\usepackage{capt-of}
\usepackage{pifont}
\usepackage{tikz}
\usepackage{enumitem}
\newtcolorbox{hypothesisbox}{
    colback=gray!8,
    colframe=black!75,
    boxrule=0.6pt,
    arc=2mm,
    left=2mm,
    right=2mm,
    top=1.5mm,
    bottom=1.5mm,
    before skip=6pt,
    after skip=6pt
}

\title{DFlow: Enabling Verifier Information Flow in Block Diffusion Speculative Decoding}

\author{
Yaojie Zhang$^{1,3}$ \quad
Linfeng Zhang$^{4}$ \quad
Bin Cui$^{1,2}$ \quad
Xupeng Miao$^{1}$\thanks{Corresponding author.} \\[3pt]
{\normalfont\fontsize{9}{10.5}\selectfont
\parbox{0.95\textwidth}{
$^{1}$School of Computer Science \& Beijing Key Laboratory of Software and Hardware Cooperative \\
\phantom{$^{1}$}Artificial Intelligence Systems, Peking University \\
$^{2}$Institute of Computational Social Science, Peking University (Qingdao) \\
$^{3}$University of Electronic Science and Technology of China \\
$^{4}$Shanghai Jiao Tong University
}}
}

\iclrfinalcopy 
\begin{document}

\maketitle
\begin{abstract}
Block diffusion speculative decoding improves LLM inference efficiency by proposing a block of future tokens in parallel and verifying them with a single forward pass through the target model. However, existing methods retain only the accepted prefix and discard the rejected suffix, preventing the computation spent on these positions from benefiting subsequent drafting rounds and forcing the drafter to repeatedly reconstruct representations for future tokens from scratch. We observe that rejection only determines whether a proposed token can be committed, while the verifier representations at rejected positions can still provide useful information for subsequent predictions. Based on this observation, we propose DFlow, a simple yet effective framework that enables verifier information to flow across drafting rounds. DFlow reuses the hidden states produced by the target verifier for the rejected suffix to guide subsequent drafting without additional target computation. To effectively learn this information flow across drafting rounds, we introduce a self-condition train strategy that feeds verifier representations from earlier predictions back into subsequent predictions. Experiments on Qwen3 models across diverse benchmarks demonstrate that DFlow consistently improves draft quality and acceptance length over DFlash.
\end{abstract}

\section{Introduction} \label{sec:intro}
Large language models (LLMs) have achieved remarkable performance across a wide range of tasks~\citep{singh2025openai,xu2026deepseek,team2026kimi}, but autoregressive decoding remains inherently sequential and often becomes a major bottleneck in LLM inference~\citep{miao2025towards,nie2026large}. 
Speculative decoding alleviates this bottleneck by using a lightweight drafter to propose multiple future tokens and verifying them in parallel with the target model~\citep{leviathan2023fast,ada}. 
Among recent approaches, block diffusion speculative decoding is particularly attractive because it predicts an entire block of future tokens simultaneously~\citep{sandler2025specdiff,liu2026dart}, substantially reducing the sequential overhead of drafting while preserving exact target model outputs through verification. 
Despite this efficiency, its practical speedup still depends heavily on how many proposed tokens can be accepted in each drafting round~\citep{zhang2026dflare,wu2026d}, making draft quality a key factor in decoding acceleration.

However, existing block diffusion speculative decoding methods impose an information discontinuity across consecutive drafting rounds. 
Given a verified draft block, only the accepted prefix is committed as context for the next drafting round, while the remaining suffix and its intermediate representations are discarded~\citep{chen2026dflash,huang2026domino}.
In the next drafting round, the drafter predicts another block of future positions, some of which overlap with positions already predicted and processed in the previous round. Nevertheless, these overlapping positions are reset to mask embeddings and reconstructed from scratch, preventing the intermediate information obtained earlier from being carried forward.
As a result, the rejection boundary serves not only as a necessary commit boundary for lossless speculative decoding, but also as an unnecessary information boundary that blocks information flow across consecutive drafting rounds.

Recent work suggests that speculative execution provides value beyond the tokens that are eventually accepted. Drafting and verification naturally produce rich intermediate artifacts, including future token predictions, attention patterns, routing signals, and contextual representations, which have increasingly been reused to anticipate KV access, prefetch experts, or reduce subsequent computation~\citep{liang2026dualdecoder,chen2025sp,chen2026make}. These approaches reveal a broader principle: a speculative proposal is not merely a candidate output, but can also serve as a probe that exposes useful information throughout the inference process. This perspective naturally extends to target verification. Rejection only determines whether a proposed token can be committed; it does not imply that the verifier representation computed at that position is uninformative. A verifier hidden state is a continuous contextual representation produced by the target model and can retain useful information about the continuation even when the corresponding discrete proposal is rejected. This is particularly relevant to the rejected suffix, whose positions have already been processed by the target model before the verification outcome is finalized. Therefore, discarding these representations together with the rejected tokens may unnecessarily erase information that could benefit subsequent drafting. This motivates a simple question: can the verifier information already computed for rejected positions be carried forward to improve the next drafting round?

To this end, we propose DFlow, a framework that enables verifier information to flow across drafting rounds in block diffusion speculative decoding. Rather than predicting overlapping future positions again from scratch, DFlow carries forward the verifier hidden states associated with the rejected suffix and uses them to guide subsequent drafting. These representations are obtained from the existing verification process and therefore introduce no additional target model forward pass.
To support this cross-round inference pattern, we further introduce Multi-Round Self-Conditioned Training, which explicitly unfolds consecutive drafting and verification rounds, allowing the training procedure to reproduce the verifier information flow that arises during DFlow inference.
As shown in Figure~\ref{fig:speed}, DFlow consistently improves draft quality over DFlash across diverse reasoning, coding, and dialogue tasks on Qwen3-1.7B, demonstrating the effectiveness of reusing verifier information beyond the rejection boundary.Our contributions are summarized as follows:

\begin{figure*}[t]
  \centering
  \includegraphics[width=1.0\linewidth]{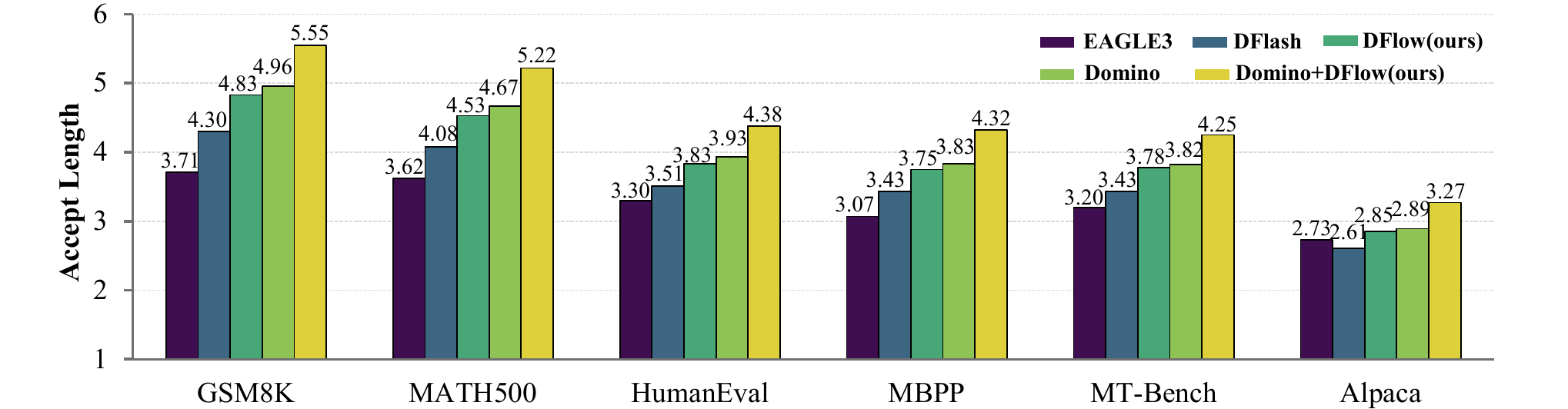}
  \caption{Average acceptance length of different methods on Qwen3-1.7B under greedy decoding.}
  \label{fig:speed}
\end{figure*}



\begin{enumerate}[leftmargin=*, topsep=0pt, itemsep=1pt, partopsep=1pt, parsep=1pt]
    \item We identify an overlooked information discontinuity in block diffusion speculative decoding:
    while the rejection boundary determines token commitment, verifier representations beyond this boundary can still provide useful guidance for subsequent drafting.

    \item We propose \textbf{DFlow}, a simple and effective framework that enables verifier information flow across drafting rounds by relaying rejected-suffix representations to corresponding positions in the next round, with Multi-Round Self-Conditioned Training to learn this information reuse.

    \item We validate DFlow on Qwen3-1.7B, Qwen3-4B, and Qwen3-8B across diverse reasoning, coding, and dialogue benchmarks. Under greedy decoding, DFlow improves the average acceptance length over DFlash by \textbf{10.4\%}, \textbf{13.2\%}, and \textbf{13.4\%} on the three model scales, respectively.

\end{enumerate}

\section{Related Work} \label{sec:related}
\textbf{Speculative Decoding for Efficient LLM Inference.}
Speculative decoding addresses the memory bound bottleneck of LLM decoding by amortizing target model memory accesses across multiple candidate tokens through parallel verification in a single forward pass~\citep{chen2023accelerating,miao2023specinfer,zhang2026flexdraft}. 
A representative line of work is the EAGLE series~\citep{li2024eagle,li2024eagle2,li2025eagle3}, which performs autoregressive speculative drafting conditioned on target model features.
To reduce the sequential overhead of such autoregressive drafting, DFlash introduces block diffusion speculative decoding, generating an entire draft block in parallel with a single draft model forward pass~\citep{chen2026dflash}. 
As drafting latency decreases, draft quality becomes the primary bottleneck to further speedup, motivating recent efforts to strengthen block parallel drafters~\citep{yang2026spec}.
DFlare improves the use of target model representations by allowing each draft layer to learn its own combination of features from multiple target layers~\citep{zhang2026dflare}, which provides more suitable information for different draft depths and enables deeper draft models to scale effectively. 
Domino focuses on the weakened causal dependency modeling within a draft block and introduces a lightweight causal head to refine parallel predictions~\citep{huang2026domino}. DSpark and DFly further explore this direction and validate such dependency modeling mechanisms under practical serving workloads~\citep{cheng2026dspark,liu2026angelspec}. 
More recently, DFlash~2 improves local dependency modeling with lightweight dynamic convolutions while preserving fully parallel drafting~\citep{inco2026dflash2}. 
Together, these methods improve draft quality and model capacity while maintaining the low drafting cost of block diffusion speculative decoding.

\textbf{Intermediate Information Reuse in Speculative Decoding.}
Recent work has increasingly explored how intermediate information produced during speculative decoding can be reused beyond its immediate role in token generation or verification.
On the draft side, speculative predictions, routing decisions, and hidden states have been exploited to anticipate KV access, prefetch experts, or support subsequent speculation~\citep{liang2026dualdecoder,li2025speculative,chen2026make}.
On the verification side, attention patterns and KV importance exposed during target verification have been reused to guide sparse attention and context selection in later drafting steps~\citep{zhao2026accelerating}.
However, existing reuse mechanisms mainly exploit intermediate artifacts to
precompute or selectively perform subsequent operations, thereby reducing
inference overhead. 
In contrast, DFlow reuses the contextual information
induced by draft proposals during target verification to directly improve
draft quality. 
Specifically, it propagates the corresponding verifier
representations beyond the rejection boundary as guidance for the next
drafting round, improving subsequent draft prediction.

\section{Preliminaries} \label{sec:preliminaries}
To facilitate the understanding of DFlow, we first review the inference process of DFlash and introduce the notation used in this paper. DFlash performs generation through a sequence of speculative rounds, each consisting of block diffusion drafting and exact-match verification.

\textbf{Block Diffusion Drafting.}
Let $\mathbf{y}_{1:t_r}$ denote the committed context at speculative round $r$, and let $K$ denote the number of future tokens drafted in parallel.
DFlash conditions a lightweight block diffusion drafter on hidden representations extracted from the target model.
Specifically, let $\mathcal{L}=\{\ell_1,\ldots,\ell_m\}$ denote a fixed set of target layers sampled from shallow to deep.
The hidden states from these layers are concatenated and projected into a fused target context representation:
\begin{equation}
\mathbf{H}^{(r)}_{T}
=
\operatorname{RMSNorm}
\left(
\mathbf{W}_{c}
\left[
\mathbf{H}^{(r,\ell_1)}_{T};
\ldots;
\mathbf{H}^{(r,\ell_m)}_{T}
\right]
\right),
\label{eq:dflash_context}
\end{equation}
where $\mathbf{H}^{(r,\ell)}_{T}$ denotes the hidden states extracted from target layer $\ell$, and $\mathbf{W}_{c}$ denotes the projection matrix.
The fused representation $\mathbf{H}^{(r)}_{T}$ is then injected into the Key and Value projections of each draft layer, providing persistent conditioning from the committed context.

To predict the next $K$ tokens, DFlash initializes the future positions with mask embeddings and jointly predicts the draft block conditioned on the fused target context representation:
\begin{equation}
\hat{\mathbf{y}}^{(r)}
=
\mathcal{D}_{\theta}
\left(
\mathbf{E}_{\mathrm{mask}}^{(r)};
\mathbf{H}_{T}^{(r)}
\right),
\label{eq:dflash_drafting}
\end{equation}
where $\mathcal{D}_{\theta}$ denotes the block diffusion drafter and $\mathbf{E}_{\mathrm{mask}}^{(r)}$ denotes the mask representations for the $K$ future positions.
While $\mathbf{H}_{T}^{(r)}$ provides target information from the committed context, the future positions are initialized from mask embedding at each speculative round, even when some of them overlap with positions already drafted and processed in the preceding round.

\textbf{Exact Match Verification.}
Given the draft block $\hat{\mathbf{y}}^{(r)}$, the target model verifies the proposed tokens from left to right using exact token matching.
Let $a_r$ denote the number of consecutive proposals that match the corresponding target predictions before the first mismatch.
Accordingly, the verified draft block can be partitioned as
\begin{equation}
\hat{\mathbf{y}}^{(r)}
=
\left[
\underbrace{\hat{\mathbf{y}}^{(r)}_{1:a_r}}_{\text{accepted prefix}},
\underbrace{\hat{y}^{(r)}_{a_r+1}}_{\text{first rejection}},
\underbrace{\hat{\mathbf{y}}^{(r)}_{a_r+2:K}}_{\text{rejected suffix}}
\right].
\label{eq:verification_partition}
\end{equation}
The first $a_r$ proposals are committed as the \emph{accepted prefix}.
At the first rejected position, the draft proposal is replaced by the corresponding target prediction, denoted by $b^{(r)}$ and referred to as the \emph{bonus token}.
The remaining proposals, referred to as the \emph{rejected suffix}, are discarded after verification.
The rejected suffix may overlap with the beginning of the subsequent drafting window, so some future positions are drafted from scratch again in the next round. Yet the information obtained at these positions is not retained across rounds, introducing an information discontinuity.

\section{DFlow} \label{sec:method}
DFlow addresses the information discontinuity across speculative rounds by preserving verifier representations associated with rejected draft positions and propagating them into subsequent drafting.
During inference, the verifier representations are propagated to their corresponding future positions and incorporated into the input of the subsequent drafting round.
To accommodate this information flow across multiple drafting rounds, we further introduce a self-conditioning train strategy that trains the drafter under both standard mask prediction and prediction with verifier information.

\subsection{Information Discontinuity across Drafting Rounds}
Existing block diffusion speculative decoding suffers from an information discontinuity across drafting rounds.
During exact match verification, only the accepted prefix and the bonus token are committed, whereas the rejected suffix is discarded.
However, positions in the rejected suffix remain unresolved after the current round and may need to be predicted again as generation proceeds.
Although the target model has already processed these positions and produced contextual representations during verification, such information is discarded together with the rejected proposals.
Consequently, the rejection boundary not only determines which tokens can be committed, but also becomes an information boundary between consecutive drafting rounds.

Importantly, rejection is a decision over discrete token commitment rather than a direct measure of information utility.
Notably, tokens beyond the first rejection can still contain useful predictive signals, with some positions remaining consistent with the target continuation despite belonging to the rejected suffix.
More importantly, during verification, each draft proposal participates in the target model's causal computation, inducing a contextual representation that integrates information from the preceding context and previously proposed tokens.
Such representations are directly involved in predicting subsequent tokens and may therefore retain useful information even when the corresponding discrete proposals are rejected. This motivates the following hypothesis:

\begin{hypothesisbox}
\textbf{Main Hypothesis.}
Verifier representations at rejected positions can provide useful guidance for subsequent drafting, improving draft quality by preserving information that would otherwise be discarded.
\end{hypothesisbox}

\begin{figure*}[t]
  \centering
  \includegraphics[width=1.0\linewidth]{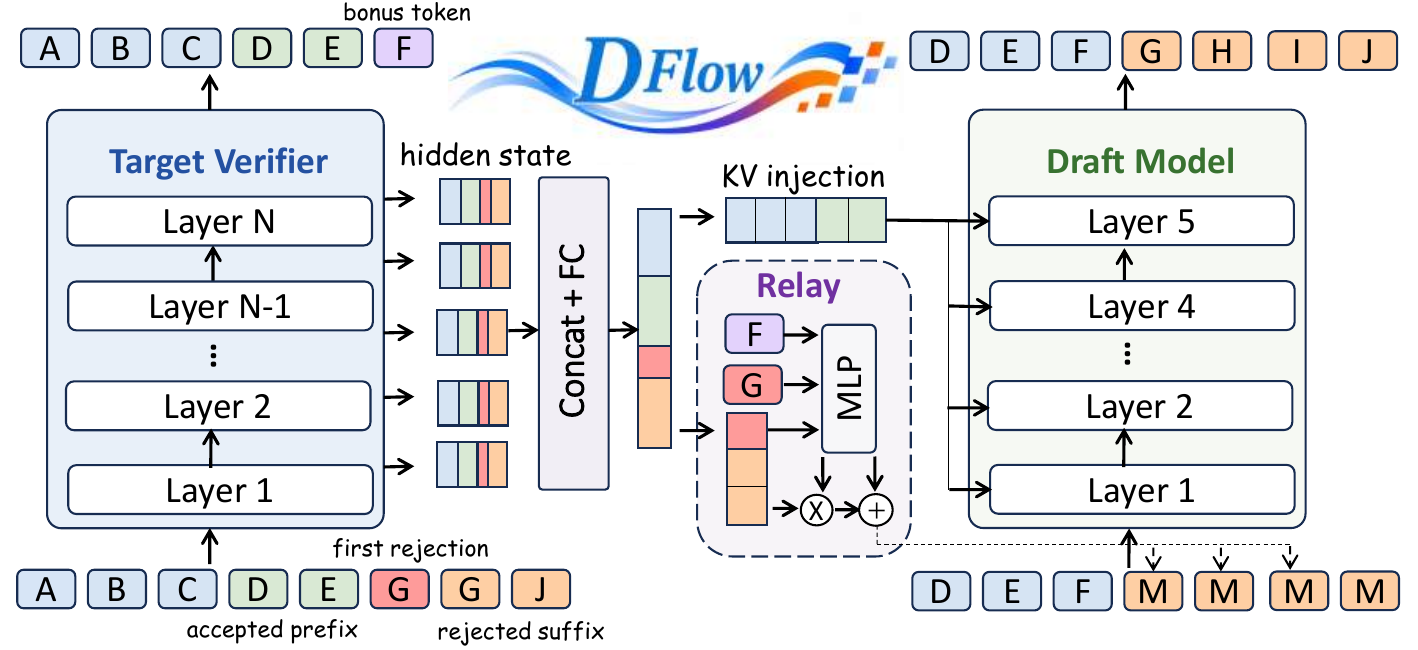}
  \caption{\textbf{Pipeline of DFlow.} DFlow reuses verifier hidden states from rejected suffix positions through the relay module to enhance the corresponding mask embeddings in the next drafting round, without an additional target model forward.
  }
  \label{fig:pipe_figure}
\end{figure*}

\subsection{Verifier Information Relay}
Based on the above hypothesis, DFlow relays verifier representations associated with the rejected suffix into subsequent drafting rounds and reuses them at the corresponding overlapping positions as auxiliary information, allowing these positions to inherit information from previous verification rather than reconstructing their representations from scratch in each round.
To make the relayed information compatible with subsequent drafting, DFlow first transforms verifier states into the drafter hidden space while maintaining their original prediction correspondence, and then refines them according to the semantic change introduced at the rejection boundary.

\textbf{Verifier Representation Alignment.}
To make the verifier hidden states compatible with the drafter hidden space, DFlow shares the same feature projection used by DFlash for target context conditioning.
Specifically, hidden states from the selected target layers over the rejected suffix are concatenated and transformed by the shared projection matrix $\mathbf{W}_c$ followed by RMSNorm, yielding verifier representations $\mathbf{h}^{(r)}_i$ for subsequent reuse.
Since target verification already produces the required hidden states, DFlow requires no additional target model forward pass.
Moreover, the reusable suffix states are jointly projected with the target features required by DFlash, introducing only a small number of additional positions.

The target verifier and the block diffusion drafter further differ in how their representations correspond to predicted tokens.
The causal target model performs next token prediction, where the hidden state at position $i$ predicts the token at position $i+1$, whereas the drafter performs mask token prediction, where each masked position predicts the token at the same position.
To align verifier representations with the drafter, DFlow shifts the verifier representations before injecting them into the drafter, so that both representations correspond to the same predicted token.
The aligned representations are then injected into the corresponding unresolved positions of the subsequent draft.

\textbf{Reject Conditioned Draft Initialization.}
Although the aligned verifier representations contain useful information about the rejected suffix, they are produced under a speculative context that differs from the context used in the subsequent drafting round.
During verification, the rejected suffix is processed following the first rejected proposal $\hat{y}^{(r)}_{a_r+1}$, whereas the next round continues from the target correction $b^{(r)}$ at the same position.
The difference between these two boundary tokens therefore provides a direct signal of how the context underlying the verifier representations changes across rounds.

DFlow uses this boundary information to determine how the verifier representation at each reusable position should be retained and corrected.
Specifically, we combine the embeddings of the target correction and the first rejected proposal with $\mathbf{h}^{(r)}_i$, and use a lightweight MLP to produce channel-wise scaling and correction vectors:
\begin{equation}
\left[
\boldsymbol{\beta}^{(r)}_i;
\boldsymbol{\gamma}^{(r)}_i
\right]
=
\operatorname{MLP}_{\phi}
\left(
\left[
\mathbf{E}\!\left(b^{(r)}\right);
\mathbf{E}\!\left(\hat{y}^{(r)}_{a_r+1}\right);
\mathbf{h}^{(r)}_i
\right]
\right),
\label{eq:reject_conditioning}
\end{equation}
where $\boldsymbol{\beta}^{(r)}_i$ controls the contribution of the verifier representation and $\boldsymbol{\gamma}^{(r)}_i$ provides an additive correction.
This channel-wise modulation allows DFlow to selectively preserve, suppress, or adjust the relayed information according to the change at the rejection boundary.

The verifier representation is first normalized and then adjusted by the learned scaling and correction vectors before being added to the standard mask embedding, forming the input for the subsequent drafting round:
\begin{equation}
\mathbf{x}^{(r+1)}_i
=
\mathbf{E}_{\mathrm{mask}}
+
\boldsymbol{\beta}^{(r)}_i
\odot
\operatorname{LN}
\left(
\mathbf{h}^{(r)}_i
\right)
+
\boldsymbol{\gamma}^{(r)}_i,
\label{eq:dflow_input}
\end{equation}
where $\operatorname{LN}(\cdot)$ denotes LayerNorm.
This augmented input is used only at positions with valid reusable verifier representations, while the remaining positions continue to use the standard mask embedding.
In the first drafting round, where no verifier information is available, or when the previous verification produces no reusable rejected suffix, DFlow simply uses mask embeddings as the draft inputs.

\subsection{Multi-Round Self-Conditioned Training}
\label{sec:self_conditioned_training}
Standard DFlash training treats each sampled drafting round independently and initializes every new round from mask embedding, whereas DFlow inference conditions later drafts on verifier representations produced in preceding rounds.
To bridge this gap, DFlow introduces Multi-Round Self-Conditioned Training, consisting of a Multi-Round Rollout that explicitly unfolds consecutive drafting and verification rounds to propagate verifier information, and a Trajectory-Guided Transition that determines the starting position of each subsequent round by matching draft proposals with the training trajectory.

\textbf{Multi-Round Rollout.}
Following DFlash, we sample anchor positions from trajectories generated by the target model and construct multiple draft blocks that can be packed for parallel training.
For each sampled anchor, DFlow extends the original single round training into a sequence of consecutive drafting rounds.
The initial round follows standard DFlash training and predicts the draft block from mask embeddings, preserving the fundamental mask prediction capability of the drafter while producing the proposals required for subsequent verification.
These proposals are processed by the frozen target model to obtain verifier representations over the rejected suffix, which are then relayed to the next round.
Subsequent rounds perform reject conditioned prediction, allowing the drafter to learn how to exploit verifier information from the preceding verification rather than relying solely on mask embeddings.
The reject conditioned stage can continue for multiple rounds, with the rollout depth $k$ controlling how many such rounds are included during training.
All unfolded rounds share the same drafter parameters, and their prediction losses jointly optimize this shared parameter set.
In practice, we unfold $k=2$ reject conditioned rounds after the initial mask prediction.

\textbf{Target Trajectory Transition.}
Each transition in the multi-round rollout requires determining both where the next drafting round begins and what verifier information is carried forward from the current round.
The starting position of the next round depends on the acceptance result of the current draft, with the new round anchor placed at the bonus token corresponding to the first mismatch.
Although target verification can directly determine the accepted prefix and bonus token, its verification result during training may occasionally differ from the precomputed target trajectory due to sampling or numerical differences.
We therefore determine the round transition by exact matching the draft proposals against the target trajectory and use the target token at the first mismatch as the bonus token.
This keeps subsequent rounds aligned with the same target trajectory, allowing the target context computed for each training sample to be directly reused while verification in later rounds only processes the newly proposed block.
Importantly, the target trajectory is used only to determine the transition between consecutive training rounds.
The verifier information carried into the next round remains tied to the actual draft proposals produced in the current round.
These proposals are processed by the frozen target model, and the verifier representations associated with the rejected suffix are extracted from this verification pass for subsequent reject conditioned prediction.

Following DFlash, we apply an exponential decay to the cross entropy loss as the draft position moves farther from the anchor, assigning greater importance to earlier predictions within each block.
The overall training objective sums the loss of the initial mask prediction and those of all subsequent reject conditioned rounds:
\begin{equation}
\mathcal{L}_{\mathrm{DFlow}}
=
\mathcal{L}^{(0)}
+
\sum_{q=1}^{k}
\mathcal{L}^{(q)},
\label{eq:dflow_training_objective}
\end{equation}
where $k$ denotes the rollout depth, $\mathcal{L}^{(0)}$ is the loss of the initial mask prediction, and $\mathcal{L}^{(q)}$ for $q \geq 1$ denotes the loss of the $q$-th reject conditioned round.
The losses from all rounds are summed before backpropagation and jointly optimize the same shared trainable drafter parameters.
The target model remains frozen throughout training, while the verifier representations and round transitions are detached from the computation graph, preventing gradients from propagating through target verification or across transitions between consecutive rounds.

\section{Experiments} \label{sec:exp}


\subsection{Experimental Setup}
\label{sec:experimental_setup}

\textbf{Models and Evaluation.}
We evaluate DFlow on Qwen3-1.7B, Qwen3-4B, and Qwen3-8B across three representative task categories: mathematical reasoning, code generation, and open-ended dialogue.
For mathematical reasoning, we use GSM8K~\citep{cobbe2021training}, MATH-500~\citep{hendrycks2021measuring}; for code generation, we evaluate on HumanEval~\citep{chen2021evaluating}, MBPP~\citep{austin2021program}; and for dialogue, we use MT-Bench~\citep{zheng2023judging} and Alpaca~\citep{taori2023alpaca}.We compare DFlow with vanilla autoregressive decoding and representative speculative decoding methods, including EAGLE-3~\citep{li2025eagle3} and DFlash~\citep{chen2026dflash}.


\textbf{Training and Implementation Details.}
We train a DFlash drafter with five layers and a draft block size of 16 from scratch for all target models.
We use 80K samples from ShareGPT\footnote{\url{https://huggingface.co/datasets/anon8231489123/ShareGPT_Vicuna_unfiltered}} for training.
Unless otherwise specified, Multi-Round Self-Conditioned Training uses $k=2$ reject conditioned rounds following the initial mask prediction.
All experiments are conducted on a single node with eight NVIDIA A800 GPUs.

\subsection{Main Results}
\label{sec:main_results}

\begin{table*}[t]
\centering
\caption{
Average acceptance length $\tau$ under the low-concurrency setting.
Results are reported under both greedy decoding ($T=0$) and sampling decoding ($T=1$).
The average is computed over all evaluated benchmarks.
}
\label{tab:main_results}

\small
\setlength{\tabcolsep}{4.5pt}
\renewcommand{\arraystretch}{1.03}

\begin{tabular}{@{}llccccccc@{}}
\toprule
& &
\multicolumn{2}{c}{\textbf{Math}} &
\multicolumn{2}{c}{\textbf{Code}} &
\multicolumn{2}{c}{\textbf{Chat}} &
\textbf{Overall} \\
\cmidrule(lr){3-4}
\cmidrule(lr){5-6}
\cmidrule(lr){7-8}
\cmidrule(l){9-9}

\textbf{Model}
& \textbf{Method}
& \textbf{GSM8K}
& \textbf{MATH-500}
& \textbf{HumanEval}
& \textbf{MBPP}
& \textbf{MT-Bench}
& \textbf{Alpaca}
& \textbf{Avg.} \\

\midrule
\multicolumn{9}{@{}l}{\textit{Temperature = 0}} \\
\midrule

\multirow{3}{*}{Qwen3-1.7B}
& EAGLE-3
& 3.71
& 3.62
& 3.30
& 3.07
& 3.20
& 2.73
& 3.27 \\

& DFlash
& 4.30
& 4.08
& 3.51
& 3.43
& 3.43
& 2.61
& 3.56 \\

& DFlow
& \textbf{4.83}
& \textbf{4.53}
& \textbf{3.83}
& \textbf{3.75}
& \textbf{3.78}
& \textbf{2.85}
& \textbf{3.93} \\

\midrule

\multirow{3}{*}{Qwen3-4B}
& EAGLE-3
& 3.30
& 3.14
& 3.08
& 3.00
& 3.07
& 2.86
& 3.08 \\

& DFlash
& 4.41
& 4.39
& 3.86
& 3.61
& 3.39
& 2.61
& 3.71 \\

& DFlow
& \textbf{5.13}
& \textbf{4.92}
& \textbf{4.33}
& \textbf{4.12}
& \textbf{3.80}
& \textbf{2.90}
& \textbf{4.20} \\

\midrule

\multirow{3}{*}{Qwen3-8B}
& EAGLE-3
& 3.25
& 3.12
& 3.25
& 2.86
& 2.88
& 2.61
& 3.00 \\

& DFlash
& 4.40
& 4.36
& 3.82
& 3.55
& 3.21
& 2.61
& 3.66 \\

& DFlow
& \textbf{5.10}
& \textbf{4.94}
& \textbf{4.28}
& \textbf{4.08}
& \textbf{3.58}
& \textbf{2.93}
& \textbf{4.15} \\

\midrule
\multicolumn{9}{@{}l}{\textit{Temperature = 1}} \\
\midrule

\multirow{3}{*}{Qwen3-1.7B}
& EAGLE-3
& 3.55
& 3.45
& 3.20
& 3.08
& 3.03
& \textbf{2.62}
& 3.16 \\

& DFlash
& 3.66
& 3.34
& 2.96
& 2.87
& 2.82
& 2.27
& 2.99 \\

& DFlow
& \textbf{4.04}
& \textbf{3.65}
& \textbf{3.21}
& \textbf{3.17}
& \textbf{3.04}
& 2.42
& \textbf{3.26} \\

\midrule

\multirow{3}{*}{Qwen3-4B}
& EAGLE-3
& 3.24
& 3.01
& 3.02
& 2.95
& 2.97
& \textbf{2.77}
& 2.99 \\

& DFlash
& 3.84
& 3.53
& 3.33
& 3.12
& 2.77
& 2.32
& 3.15 \\

& DFlow
& \textbf{4.40}
& \textbf{3.97}
& \textbf{3.70}
& \textbf{3.55}
& \textbf{3.13}
& 2.49
& \textbf{3.54} \\

\midrule

\multirow{3}{*}{Qwen3-8B}
& EAGLE-3
& 3.15
& 2.94
& 3.07
& 2.80
& 2.73
& 2.48
& 2.86 \\

& DFlash
& 3.84
& 3.54
& 3.21
& 3.06
& 2.67
& 2.30
& 3.10 \\

& DFlow
& \textbf{4.31}
& \textbf{3.96}
& \textbf{3.58}
& \textbf{3.44}
& \textbf{2.92}
& \textbf{2.50}
& \textbf{3.45} \\

\bottomrule
\end{tabular}
\end{table*}

\textbf{Low Concurrency.}
Table~\ref{tab:main_results} reports the average acceptance length $\tau$ across all evaluated benchmarks under both greedy and sampling decoding.
Across different model scales and task categories, DFlow consistently improves upon DFlash, demonstrating that verifier information from rejected suffixes provides useful guidance for subsequent drafting.
Under greedy decoding ($T=0$), DFlow increases the average acceptance length from
\textbf{3.56} to \textbf{3.93},
\textbf{3.71} to \textbf{4.20}, and
\textbf{3.66} to \textbf{4.15}
on Qwen3-1.7B, Qwen3-4B, and Qwen3-8B, respectively.
Similar improvements are observed under sampling decoding ($T=1$), where the average acceptance length increases from
\textbf{2.99} to \textbf{3.26},
\textbf{3.15} to \textbf{3.54}, and
\textbf{3.10} to \textbf{3.45}.
These consistent gains across reasoning, coding, and dialogue benchmarks show that DFlow effectively improves draft quality across different model scales and decoding settings.

\begin{table}[t]
\centering
\caption{
High-concurrency throughput on Qwen3 models.
Baseline rows report absolute throughput in TPS.
Other entries report TPS with green subscripts indicating speedup over the corresponding baseline.
}
\label{tab:high_concurrency}

\small
\setlength{\tabcolsep}{3.6pt}
\renewcommand{\arraystretch}{0.92}

\resizebox{\linewidth}{!}{%
\begin{tabular}{llllllll}
\toprule
\textbf{Task} & \textbf{Method}
& \multicolumn{6}{c}{\textbf{Concurrency}} \\
\cmidrule(lr){3-8}
&
& \multicolumn{1}{c}{1}
& \multicolumn{1}{c}{2}
& \multicolumn{1}{c}{4}
& \multicolumn{1}{c}{8}
& \multicolumn{1}{c}{16}
& \multicolumn{1}{c}{32} \\
\midrule

\multicolumn{8}{l}{\textbf{Qwen3-4B}} \\
\midrule

\multirow{4}{*}{GSM8K}
& Baseline
& 146.25
& 287.58
& 549.67
& 1063.16
& 1804.22
& 2850.74 \\

& EAGLE-3
& 233.99$_{\textcolor{green!50!black}{1.60\times}}$
& 420.16$_{\textcolor{green!50!black}{1.46\times}}$
& 758.99$_{\textcolor{green!50!black}{1.38\times}}$
& 1210.98$_{\textcolor{green!50!black}{1.14\times}}$
& 1634.96$_{\textcolor{green!50!black}{0.91\times}}$
& 1942.57$_{\textcolor{green!50!black}{0.68\times}}$ \\

& DFlash
& 349.59$_{\textcolor{green!50!black}{2.39\times}}$
& 635.08$_{\textcolor{green!50!black}{2.21\times}}$
& 1124.58$_{\textcolor{green!50!black}{2.05\times}}$
& 1802.80$_{\textcolor{green!50!black}{1.70\times}}$
& 2368.26$_{\textcolor{green!50!black}{1.31\times}}$
& 2672.69$_{\textcolor{green!50!black}{0.94\times}}$ \\

& DFlow
& \textbf{376.26}$_{\textcolor{green!50!black}{2.57\times}}$
& \textbf{679.39}$_{\textcolor{green!50!black}{2.36\times}}$
& \textbf{1217.36}$_{\textcolor{green!50!black}{2.21\times}}$
& \textbf{1959.07}$_{\textcolor{green!50!black}{1.84\times}}$
& \textbf{2580.66}$_{\textcolor{green!50!black}{1.43\times}}$
& \textbf{3208.95}$_{\textcolor{green!50!black}{1.13\times}}$ \\

\cmidrule(lr){2-8}

\multirow{4}{*}{HumanEval}
& Baseline
& 145.58
& 287.16
& 544.38
& 1015.14
& 1742.29
& 2427.99 \\

& EAGLE-3
& 220.83$_{\textcolor{green!50!black}{1.52\times}}$
& 398.01$_{\textcolor{green!50!black}{1.39\times}}$
& 742.55$_{\textcolor{green!50!black}{1.36\times}}$
& 1201.58$_{\textcolor{green!50!black}{1.18\times}}$
& 1599.72$_{\textcolor{green!50!black}{0.92\times}}$
& 1888.08$_{\textcolor{green!50!black}{0.78\times}}$ \\

& DFlash
& 334.06$_{\textcolor{green!50!black}{2.29\times}}$
& 604.59$_{\textcolor{green!50!black}{2.11\times}}$
& 1066.61$_{\textcolor{green!50!black}{1.96\times}}$
& 1750.33$_{\textcolor{green!50!black}{1.72\times}}$
& 2336.69$_{\textcolor{green!50!black}{1.34\times}}$
& 2835.77$_{\textcolor{green!50!black}{1.17\times}}$ \\

& DFlow
& \textbf{343.42}$_{\textcolor{green!50!black}{2.36\times}}$
& \textbf{617.27}$_{\textcolor{green!50!black}{2.15\times}}$
& \textbf{1107.12}$_{\textcolor{green!50!black}{2.03\times}}$
& \textbf{1850.42}$_{\textcolor{green!50!black}{1.82\times}}$
& \textbf{2482.46}$_{\textcolor{green!50!black}{1.42\times}}$
& \textbf{3089.40}$_{\textcolor{green!50!black}{1.27\times}}$ \\

\midrule

\multicolumn{8}{l}{\textbf{Qwen3-8B}} \\
\midrule

\multirow{4}{*}{GSM8K}
& Baseline
& 91.91
& 181.31
& 356.37
& 662.73
& 1070.40
& 1630.10 \\

& EAGLE-3
& 170.03$_{\textcolor{green!50!black}{1.85\times}}$
& 310.70$_{\textcolor{green!50!black}{1.71\times}}$
& 545.32$_{\textcolor{green!50!black}{1.53\times}}$
& 930.10$_{\textcolor{green!50!black}{1.40\times}}$
& 1244.19$_{\textcolor{green!50!black}{1.16\times}}$
& 1396.23$_{\textcolor{green!50!black}{0.86\times}}$ \\

& DFlash
& 260.26$_{\textcolor{green!50!black}{2.83\times}}$
& 473.12$_{\textcolor{green!50!black}{2.61\times}}$
& 877.87$_{\textcolor{green!50!black}{2.46\times}}$
& 1404.98$_{\textcolor{green!50!black}{2.12\times}}$
& 1835.42$_{\textcolor{green!50!black}{1.71\times}}$
& 1891.10$_{\textcolor{green!50!black}{1.16\times}}$ \\

& DFlow
& \textbf{283.47}$_{\textcolor{green!50!black}{3.08\times}}$
& \textbf{508.97}$_{\textcolor{green!50!black}{2.81\times}}$
& \textbf{944.11}$_{\textcolor{green!50!black}{2.65\times}}$
& \textbf{1518.16}$_{\textcolor{green!50!black}{2.29\times}}$
& \textbf{2023.22}$_{\textcolor{green!50!black}{1.89\times}}$
& \textbf{2198.17}$_{\textcolor{green!50!black}{1.35\times}}$ \\

\cmidrule(lr){2-8}

\multirow{4}{*}{HumanEval}
& Baseline
& 91.86
& 181.98
& 352.96
& 663.87
& 1200.25
& 2036.43 \\

& EAGLE-3
& 168.83$_{\textcolor{green!50!black}{1.84\times}}$
& 306.27$_{\textcolor{green!50!black}{1.68\times}}$
& 565.86$_{\textcolor{green!50!black}{1.60\times}}$
& 947.66$_{\textcolor{green!50!black}{1.43\times}}$
& 1296.97$_{\textcolor{green!50!black}{1.08\times}}$
& 1467.19$_{\textcolor{green!50!black}{0.72\times}}$ \\

& DFlash
& 230.75$_{\textcolor{green!50!black}{2.51\times}}$
& 422.87$_{\textcolor{green!50!black}{2.32\times}}$
& 809.74$_{\textcolor{green!50!black}{2.29\times}}$
& 1296.37$_{\textcolor{green!50!black}{1.95\times}}$
& 1777.93$_{\textcolor{green!50!black}{1.48\times}}$
& 1986.28$_{\textcolor{green!50!black}{0.98\times}}$ \\

& DFlow
& \textbf{245.12}$_{\textcolor{green!50!black}{2.67\times}}$
& \textbf{448.55}$_{\textcolor{green!50!black}{2.46\times}}$
& \textbf{849.26}$_{\textcolor{green!50!black}{2.41\times}}$
& \textbf{1368.45}$_{\textcolor{green!50!black}{2.06\times}}$
& \textbf{1864.77}$_{\textcolor{green!50!black}{1.55\times}}$
& \textbf{2101.42}$_{\textcolor{green!50!black}{1.03\times}}$ \\

\bottomrule
\end{tabular}%
}
\end{table}
\textbf{High Concurrency.}
We further evaluate DFlow under concurrent serving workloads using SGLang.
Table~\ref{tab:high_concurrency} reports serving throughput on Qwen3-4B and Qwen3-8B under concurrency levels from 1 to 32.
Across both model scales and workloads, DFlow consistently achieves higher throughput than DFlash, demonstrating that improvements in draft quality translate into practical serving efficiency.
The throughput gains persist across the entire concurrency range and, in several cases, become larger under heavier serving loads.
This suggests that the additional computation introduced by verifier information relay is small relative to the benefit of accepting more draft tokens, allowing DFlow to remain effective as batching more fully utilizes the target model.

\subsection{Ablation Studies}
\label{sec:ablation}

\begin{figure*}[t]
    \centering
    \small
    \renewcommand{\arraystretch}{1.08}

    \begin{minipage}[t]{0.50\textwidth}
        \vspace{0pt}
        \centering
        \captionof{table}{Ablation of verifier information relay.}
        \label{tab:relay_ablation}

        \vspace{1pt}
        \setlength{\tabcolsep}{8pt}

        \begin{tabular}{lcc}
        \toprule
        \textbf{Training} & \textbf{Mask} & \textbf{Relay} \\
        \midrule
        DFlash                     & 3.56 & -- \\
        DFlash + Self-Conditioning & 3.67 & -- \\
        \textbf{DFlow}             & 3.66 & \textbf{3.93} \\
        \bottomrule
        \end{tabular}
    \end{minipage}
    \hfill
    \begin{minipage}[t]{0.47\textwidth}
        \vspace{10pt}
        \centering

        \includegraphics[
            width=0.94\linewidth,
            trim=4pt 2pt 4pt 2pt,
            clip
        ]{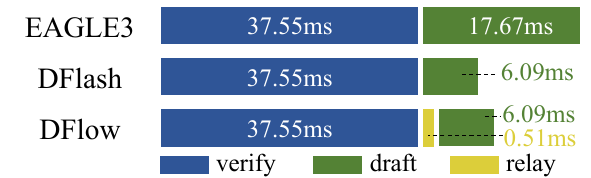}

        \vspace{-8pt}
        \captionof{figure}{Runtime overhead.}
        \label{fig:overhead}
    \end{minipage}

\end{figure*}
\textbf{Effect of Verifier Information Relay.}
To quantify the respective contributions of Multi-Round Self-Conditioned Training and verifier information relay, we conduct controlled ablations.
We first retain DFlow's multi-round rollout while resetting each new round to standard mask embeddings, thereby isolating the benefit of multi-round mask prediction from verifier information reuse.
This improves the average acceptance length from 3.56 to 3.67, indicating that multi-round training alone provides only a modest gain.
We then evaluate the trained DFlow model with verifier relay disabled, where all draft positions are initialized from the standard mask embedding.
Its acceptance length of 3.66 closely matches the 3.67 achieved by multi-round mask training, showing that DFlow builds upon a strong mask token prediction capability.
Finally, enabling verifier information relay increases the acceptance length from 3.66 to 3.93 under the same trained model.
This clear improvement confirms that verifier information reuse is the primary source of DFlow's gain.

\textbf{Runtime Overhead Analysis.}
We decompose the latency of each speculative round to quantify the additional cost introduced by DFlow.
As shown in Figure~\ref{fig:overhead}, the dominant computation remains draft model execution and target verification, while verifier information relay introduces only 0.51\,ms of additional latency, corresponding to merely \textbf{1.2\%} overhead over DFlash.
This low overhead is because the required hidden states are already produced during target verification and DFlow reuses the target feature projection of DFlash; the additional computation mainly consists of processing a small number of reusable rejected positions and applying the lightweight relay module.
Compared with the substantial improvement in acceptance length, this additional cost is negligible, allowing the improved draft quality to translate into higher end-to-end decoding throughput.


\section{Compatibility with Domino} \label{sec:discussion}
Recent studies have shown that autoregressive heads can effectively recover causal dependencies among parallel draft tokens and substantially improve draft quality~\citep{cheng2026dspark}. We therefore examine whether DFlow remains beneficial when combined with this stronger drafting paradigm, using Domino as a representative method that decouples causal dependency modeling from expensive autoregressive draft execution. Domino progressively refines predictions across positions within each drafting round~\citep{huang2026domino}, whereas DFlow propagates verifier information across rounds. To combine the two, we introduce an \emph{online rollout} during DFlow training to ensure that the rejected suffix used for relay construction follows the same draft trajectory as inference.






\begin{wraptable}[8]{r}{0.43 \columnwidth}
    \vspace{-2.2em}
    \centering
    \caption{Compatibility with Domino.}
    \label{tab:domino_rollout}
    \renewcommand{\arraystretch}{1.08}
    \begin{tabular}{l c}
    \toprule
    \textbf{Method} & $\bm{\tau}$ \\
    \midrule
    DFlash & 3.56 \\
    Domino & 4.02 \\
    + DFlow (w/o Online Rollout) & 2.22 \\
    \textbf{+ DFlow (Online Rollout)} & \textbf{4.50} \\
    \bottomrule
    \end{tabular}
\end{wraptable}

Domino is trained with teacher forcing, where each causal prediction is conditioned on the ground truth prefix and contributes to the cross entropy objective of the current round. 
Directly using these predictions to determine the rejected suffix and extract its verifier representations for DFlow, however, introduces a substantial mismatch with inference, where each position is conditioned on previously generated draft tokens. 
Teacher forcing maintains considerably higher accuracy at later draft positions, making the resulting rejected suffix overly optimistic for relay training. We therefore retain teacher forcing for optimizing Domino, while separately rolling out its actual predictions to determine the rejection boundary and extract the corresponding verifier representations for DFlow. Without online rollout, the average acceptance length drops to $2.22$. With online rollout, DFlow and Domino can be effectively combined, as shown in Table~\ref{tab:domino_rollout}. On Qwen3 1.7B at $T=0$, Domino achieves an average acceptance length of $4.02$, compared with $3.56$ for DFlash, while adding DFlow further improves it to $4.50$. This result shows that verifier information reuse remains effective when combined with stronger causal dependency modeling.


\section{Conclusion} \label{sec:conclusion}
We propose DFlow, a block diffusion speculative decoding framework that enables verifier information to flow across consecutive drafting rounds. DFlow relays verifier representations from the rejected suffix to subsequent unresolved positions and introduces Multi-Round Self-Conditioned Training to learn this information reuse, without requiring additional target model forward passes.
Experiments on Qwen3 models across diverse benchmarks show consistent gains over DFlash in acceptance length and decoding efficiency, while remaining compatible with stronger drafting methods such as Domino.
These results suggest that verifier information beyond the rejection boundary can be effectively reused to improve subsequent drafting for LLM inference acceleration.

\bibliography{iclr2026_conference}

@inproceedings{ada,
  author       = {Zikun Li and
                  Zhuofu Chen and
                  Remi Delacourt and
                  Gabriele Oliaro and
                  Zeyu Wang and
                  Qinghan Chen and
                  Shuhuai Lin and
                  April Yang and
                  Zhihao Zhang and
                  Zhuoming Chen and
                  Yi{-}Hsiang Lai and
                  Xinhao Cheng and
                  Xupeng Miao and
                  Zhihao Jia},
  editor       = {Antonio Barbalace and
                  Luo Mai and
                  Roxana Geambasu and
                  Peter R. Pietzuch},
  title        = {AdaServe: Accelerating Multi-SLO {LLM} Serving with SLO-Customized
                  Speculative Decoding},
  booktitle    = {Proceedings of the 21st European Conference on Computer Systems, EuroSys
                  2026, McEwan Hall/The University of Edinburgh, Edinburgh, Scotland,
                  UK, April 27-30, 2026},
  pages        = {36--54},
  publisher    = {{ACM}},
  year         = {2026},
  url          = {https://doi.org/10.1145/3767295.3769315},
  doi          = {10.1145/3767295.3769315},
  bibsource    = {dblp computer science bibliography, https://dblp.org}
}

@article{xu2026deepseek,
  author       = {DeepSeek{-}AI},
  title        = {DeepSeek-V4: Towards Highly Efficient Million-Token Context Intelligence},
  journal      = {CoRR},
  volume       = {abs/2606.19348},
  year         = {2026},
  url          = {https://doi.org/10.48550/arXiv.2606.19348},
  doi          = {10.48550/ARXIV.2606.19348},
  eprinttype   = {arXiv},
  eprint       = {2606.19348},
  bibsource    = {dblp computer science bibliography, https://dblp.org}
}

@article{team2026kimi, author = {{Kimi Team}}, title = {Kimi {K3:} Open Frontier Intelligence}, journal = {CoRR}, volume = {abs/2607.24653}, year = {2026}, url = {https://doi.org/10.48550/arXiv.2607.24653}, doi = {10.48550/ARXIV.2607.24653}, eprinttype = {arXiv}, eprint = {2607.24653}, bibsource = {dblp computer science bibliography, https://dblp.org} }

@article{singh2025openai,
  title={Openai gpt-5 system card},
  author={Singh, Aaditya and Fry, Adam and Perelman, Adam and Tart, Adam and Ganesh, Adi and El-Kishky, Ahmed and McLaughlin, Aidan and Low, Aiden and Ostrow, AJ and Ananthram, Akhila and others},
  journal={arXiv preprint arXiv:2601.03267},
  year={2025}
}

@inproceedings{leviathan2023fast,
  title={Fast Inference from Transformers via Speculative Decoding},
  author={Leviathan, Yaniv and Kalman, Matan and Matias, Yossi},
  booktitle={Proceedings of the 40th International Conference on Machine Learning},
  pages={19274--19286},
  year={2023},
  volume={202},
  series={Proceedings of Machine Learning Research},
  publisher={PMLR}
}

@inproceedings{li2024eagle,
  title={{EAGLE}: Speculative Sampling Requires Rethinking Feature Uncertainty},
  author={Li, Yuhui and Wei, Fangyun and Zhang, Chao and Zhang, Hongyang},
  booktitle={Proceedings of the 41st International Conference on Machine Learning},
  pages={28935--28948},
  year={2024},
  volume={235},
  series={Proceedings of Machine Learning Research},
  publisher={PMLR}
}

@inproceedings{li2024eagle2,
  title={{EAGLE}-2: Faster Inference of Language Models with Dynamic Draft Trees},
  author={Li, Yuhui and Wei, Fangyun and Zhang, Chao and Zhang, Hongyang},
  booktitle={Proceedings of the 2024 Conference on Empirical Methods in Natural Language Processing},
  pages={7421--7432},
  year={2024},
  publisher={Association for Computational Linguistics},
  doi={10.18653/v1/2024.emnlp-main.422}
}

@inproceedings{li2025eagle3,
  title={{EAGLE}-3: Scaling up Inference Acceleration of Large Language Models via Training-Time Test},
  author={Li, Yuhui and Wei, Fangyun and Zhang, Chao and Zhang, Hongyang},
  booktitle={Advances in Neural Information Processing Systems},
  year={2025}
}

@article{chen2026dflash,
  title={DFlash: Block Diffusion for Flash Speculative Decoding},
  author={Chen, Jian and Liang, Yesheng and Liu, Zhijian},
  journal={arXiv preprint arXiv:2602.06036},
  year={2026}
}

@article{liu2026dart,
  title={DART: Diffusion-Inspired Speculative Decoding for Fast LLM Inference},
  author={Liu, Fuliang and Li, Xue and Zhao, Ketai and Gao, Yinxi and Zhou, Ziyan and Zhang, Zhonghui and Wang, Zhibin and Dou, Wanchun and Zhong, Sheng and Tian, Chen},
  journal={arXiv preprint arXiv:2601.19278},
  year={2026}
}

@article{zhang2026dflare,
  title={DFlare: Scaling Up Draft Capacity for Block Diffusion Speculative Decoding},
  author={Zhang, Jiebin and Yu, Zhenghan and Liu, Song and Yu, Eugene J and Li, Zheng and Zhu, Dawei and Duo, Jiangshan and Xiong, Weimin and Song, Yifan and Yu, Guanghua and others},
  journal={arXiv preprint arXiv:2606.02091},
  year={2026}
}

@article{nie2026large,
  title={Large Language Diffusion Models},
  author={Nie, Shen and Zhu, Fengqi and You, Zebin and Zhang, Xiaolu and Ou, Jingyang and Hu, Jun and Zhou, Jun and Lin, Yankai and Wen, Ji-Rong and Li, Chongxuan},
  journal={arXiv preprint arXiv:2502.09992},
  year={2025}
}

@article{miao2025towards,
  title={Towards efficient generative large language model serving: A survey from algorithms to systems},
  author={Miao, Xupeng and Oliaro, Gabriele and Zhang, Zhihao and Cheng, Xinhao and Jin, Hongyi and Chen, Tianqi and Jia, Zhihao},
  journal={ACM Computing Surveys},
  volume={58},
  number={1},
  pages={1--37},
  year={2025},
  publisher={ACM New York, NY}
}

@article{wu2026d,
  title={D-PACE: Dynamic Position-Aware Cross-Entropy for Parallel Speculative Drafting},
  author={Wu, Tianyu and Yao, Yu and Qi, Zhenting and Zheng, Han and Wang, Zhuohan and Ma, Haoran and Liao, Lawrence and Lakkaraju, Himabindu and Li, Ju and Du, Yilun},
  journal={arXiv preprint arXiv:2605.18810},
  year={2026}
}

@article{yang2026spec,
  title={Spec-AUF: Accept-Until-Fail Training under Train-Inference Misalignment for Masked Block Drafters},
  author={Yang, Tianjian and Li, Meng},
  journal={arXiv preprint arXiv:2607.01893},
  year={2026}
}

@article{huang2026domino,
  title={Domino: Decoupling causal modeling from autoregressive drafting in speculative decoding},
  author={Huang, Jianuo and Zhang, Yaojie and Zhang, Qituan and Lin, Hao and Xu, Hanlin and Zhang, Linfeng},
  journal={arXiv preprint arXiv:2605.29707},
  year={2026}
}

@misc{inco2026dflash2,
  title  = {DFlash 2: Keep Drafting Parallel},
  author = {{Inco AI}},
  year   = {2026},
  month  = {August},
  url    = {https://inco.ai/blog/dflash2/}
}

@article{cheng2026dspark,
  title={DSpark: Confidence-Scheduled Speculative Decoding with Semi-Autoregressive Generation},
  author={Cheng, Xin and Yu, Xingkai and Shao, Chenze and Li, Jiashi and Xiong, Yunfan and Qian, Yi and Zhu, Jiaqi and Ma, Shirong and Zhang, Xiaokang and Ye, Jiasheng and others},
  journal={arXiv preprint arXiv:2607.05147},
  year={2026}
}

@article{chen2026make,
  title={Make Every Draft Count: Hidden State based Speculative Decoding},
  author={Chen, Yuetao and Wang, Xuliang and Zheng, Xinzhou and Li, Ming and Wang, Peng and Xu, Hong},
  journal={arXiv preprint arXiv:2602.21224},
  year={2026}
}

@article{zhang2026flexdraft,
  title={FlexDraft: Flexible Speculative Decoding via Attention Tuning and Bonus-Guided Calibration},
  author={Zhang, Yaojie and Huang, Jianuo and Ke, Junlong and Han, Yuhang and Long, Yongji and Zhao, Tianchen and Qi, Biqing and Zhang, Linfeng},
  journal={arXiv preprint arXiv:2605.20022},
  year={2026}
}

@article{liu2026angelspec,
  title={AngelSpec: Towards Real-World High Performance Inference with Speculative Decoding},
  author={Liu, Hong and Cen, Rui and Shi, Junhan and Qin, Guangshuo and Zhang, Jiebin and Liu, Tianyu and Fan, Runzhi and Zhao, Guoliang and Xie, Ruobing and Zhang, Kai and others},
  journal={arXiv preprint arXiv:2607.25852},
  year={2026}
}

@article{liang2026dualdecoder,
  title={DualDecoder: Accelerate Long Context LLM Inference by Predictive Prefetch},
  author={Liang, Zuning and Yao, Zhiyi and Chen, Qi and Xu, Yuedong and Dai, Hao and Ding, Zhiqiang and Yang, Tongkai and Hou, Jinlong and Cheng, Yuan},
  journal={arXiv preprint arXiv:2607.26475},
  year={2026}
}

@article{chen2025sp,
  title={SP-MoE: Speculative Decoding and Prefetching for Accelerating MoE-based Model Inference},
  author={Chen, Liangkun and Wen, Zijian and Wu, Tian and Zhang, Xiaoxi and Wu, Chuan},
  journal={arXiv preprint arXiv:2510.10302},
  year={2025}
}

@article{chen2023accelerating,
  title={Accelerating large language model decoding with speculative sampling},
  author={Chen, Charlie and Borgeaud, Sebastian and Irving, Geoffrey and Lespiau, Jean-Baptiste and Sifre, Laurent and Jumper, John},
  journal={arXiv preprint arXiv:2302.01318},
  year={2023}
}

@article{zhao2026accelerating,
  title={Accelerating Large-Scale Reasoning Model Inference with Sparse Self-Speculative Decoding},
  author={Zhao, Yilong and Tang, Jiaming and Zhu, Kan and Ye, Zihao and Chang, Chi-Chih and Lin, Chaofan and Park, Jongseok and Xiao, Guangxuan and Abdelfattah, Mohamed S and Gao, Mingyu and others},
  journal={Proceedings of Machine Learning and Systems},
  volume={8},
  pages={251--265},
  year={2026}
}

@article{li2025speculative,
  title={Speculative MoE: Communication efficient parallel moe inference with speculative token and expert pre-scheduling},
  author={Li, Yan and Zheng, Pengfei and Chen, Shuang and Xu, Zewei and Lai, Yuanhao and Du, Yunfei and Wang, Zhengang},
  journal={arXiv preprint arXiv:2503.04398},
  year={2025}
}

@article{miao2023specinfer,
  title={Specinfer: Accelerating generative large language model serving with tree-based speculative inference and verification},
  author={Miao, Xupeng and Oliaro, Gabriele and Zhang, Zhihao and Cheng, Xinhao and Wang, Zeyu and Zhang, Zhengxin and Wong, Rae Ying Yee and Zhu, Alan and Yang, Lijie and Shi, Xiaoxiang and others},
  journal={arXiv preprint arXiv:2305.09781},
  year={2023}
}

@article{sandler2025specdiff,
  title={Specdiff-2: Scaling diffusion drafter alignment for faster speculative decoding},
  author={Sandler, Jameson and Christopher, Jacob K and Hartvigsen, Thomas and Fioretto, Ferdinando},
  journal={arXiv preprint arXiv:2511.00606},
  year={2025}
}

@article{cobbe2021training,
  title={Training Verifiers to Solve Math Word Problems},
  author={Cobbe, Karl and Kosaraju, Vineet and Bavarian, Mohammad and Chen, Mark and Jun, Heewoo and Kaiser, Lukasz and Plappert, Matthias and Tworek, Jerry and Hilton, Jacob and Nakano, Reiichiro and Hesse, Christopher and Schulman, John},
  journal={arXiv preprint arXiv:2110.14168},
  year={2021}
}

@article{hendrycks2021measuring,
  title={Measuring Mathematical Problem Solving With the MATH Dataset},
  author={Hendrycks, Dan and Burns, Collin and Kadavath, Saurav and Arora, Akul and Basart, Steven and Tang, Eric and Song, Dawn and Steinhardt, Jacob},
  journal={arXiv preprint arXiv:2103.03874},
  year={2021}
}

@article{chen2021evaluating,
  title={Evaluating Large Language Models Trained on Code},
  author={Chen, Mark and Tworek, Jerry and Jun, Heewoo and Yuan, Qiming and Pinto, Henrique Ponde de Oliveira and Kaplan, Jared and Edwards, Harri and Burda, Yuri and Joseph, Nicholas and Brockman, Greg and Ray, Alex and Puri, Raul and Krueger, Gretchen and Petrov, Michael and Khlaaf, Heidy and Sastry, Girish and Mishkin, Pamela and Chan, Brooke and Gray, Scott and Ryder, Nick and Pavlov, Mikhail and Power, Alethea and Kaiser, Lukasz and Bavarian, Mohammad and Winter, Clemens and Tillet, Philippe and Such, Felipe Petroski and Cummings, Dave and Plappert, Matthias and Chantzis, Fotios and Barnes, Elizabeth and Herbert-Voss, Ariel and Guss, William Hebgen and Nichol, Alex and Paino, Alex and Tezak, Nikolas and Tang, Jie and Babuschkin, Igor and Balaji, Suchir and Jain, Shantanu and Saunders, William and Hesse, Christopher and Carr, Andrew N. and Leike, Jan and Achiam, Josh and Misra, Vedant and Morikawa, Evan and Radford, Alec and Knight, Matthew and Brundage, Miles and Murati, Mira and Mayer, Katie and Welinder, Peter and McGrew, Bob and Amodei, Dario and McCandlish, Sam and Sutskever, Ilya and Zaremba, Wojciech},
  journal={arXiv preprint arXiv:2107.03374},
  year={2021}
}

@article{austin2021program,
  title={Program Synthesis with Large Language Models},
  author={Austin, Jacob and Odena, Augustus and Nye, Maxwell and Bosma, Maarten and Michalewski, Henryk and Dohan, David and Jiang, Ellen and Cai, Carrie and Terry, Michael and Le, Quoc and Sutton, Charles},
  journal={arXiv preprint arXiv:2108.07732},
  year={2021}
}

@inproceedings{zheng2023judging,
  title={Judging LLM-as-a-Judge with MT-Bench and Chatbot Arena},
  author={Zheng, Lianmin and Chiang, Wei-Lin and Sheng, Ying and Zhuang, Siyuan and Wu, Zhanghao and Zhuang, Yonghao and Lin, Zi and Li, Zhuohan and Li, Dacheng and Xing, Eric P. and Zhang, Hao and Gonzalez, Joseph E. and Stoica, Ion},
  booktitle={Advances in Neural Information Processing Systems},
  year={2023}
}

@article{taori2023alpaca,
  title={Alpaca: A Strong, Replicable Instruction-Following Model},
  author={Taori, Rohan and Gulrajani, Ishaan and Zhang, Tianyi and Dubois, Yann and Li, Xuechen and Guestrin, Carlos and Liang, Percy and Hashimoto, Tatsunori B.},
  journal={Stanford Center for Research on Foundation Models},
  year={2023},
  note={https://crfm.stanford.edu/2023/03/13/alpaca.html}
}
\bibliographystyle{iclr2026_conference}

\newpage
\appendix
\section{Appendix}

\subsection{Implementation Details}
\label{sec:exp_details}

For EAGLE-3, we directly use the released model checkpoint provided by AngelSlim\footnote{\url{https://github.com/Tencent/AngelSlim}}.
DFlash and DFlow are both trained on ShareGPT for 6 epochs with a learning rate of $6\times10^{-4}$.
Unless otherwise specified, DFlash and DFlow use identical training data and optimization settings for a fair comparison.
For evaluation, we set the maximum number of newly generated tokens to 2048 for all methods.

\end{document}